\documentclass{article}
\usepackage{spconf,amsmath,graphicx}
\usepackage{tikz}
\usepackage{pgfplots}
\usepackage{graphicx}

\newcommand{\myfigref}[1]{Fig. \ref{#1}}
\newcommand{\mytabref}[1]{Table. \ref{#1}}

\title{SFPN: Synthetic FPN for Object Detection}
\name{$^1$Yu-Ming Zhang, $^2$Jun-Wei Hsieh, $^1$Chun-Chieh Lee, $^1$Kuo-Chin Fan}
\address{$^1$National Central University, $^2$National Yang Ming Chiao Tung University\\
         $^1$Department of Computer Science and Information Engineering,\\
         $^2$College of Artificial Intelligence and Green Energy\\
         $^1$Taoyuan, Taiwan, $^2$Hsinchu, Taiwan}

\begin{document}
%
\maketitle
\begin{abstract}
FPN (Feature Pyramid Network) has become a basic component of most SoTA one stage object detectors.  Many previous studies have repeatedly proved that FPN can caputre better multi-scale feature maps to more precisely describe objects if they are with different sizes.  However, for most backbones such VGG, ResNet, or DenseNet, the feature maps at each layer are downsized to their quarters due to the pooling operation or convolutions with stride 2. The gap of down-scaling-by-2 is large and makes its FPN not fuse the features smoothly. This paper proposes a new SFPN (Synthetic Fusion Pyramid Network) arichtecture which creates various synthetic layers between layers of the original FPN to enhance the accuracy of light-weight CNN backones to extract objects' visual features more accurately. Finally, experiments prove the SFPN architecture outperforms either the large backbone VGG16, ResNet50 or light-weight backbones such as MobilenetV2 based on AP score.
\end{abstract}

\section{Introduction}
\label{sec:intro}
Many past studies\cite{ssd,fpn,retinanet,yolov2} have shown that feature maps in a feature pyramid can capture an object's visual features at different scales. The shallow layers retain details, such as texture, corner, and so on; the deep layers cover a broader range of semantic features.  In real scenes, objects with different sizes often appear together, and how to detect them simultaneously becomes a critical problem. This way makes FPN~\cite{fpn} significantly improve object detection performance and become a standard component of most SoTA object detectors~\cite{yolov1,prb}. However,
the feaure maps in this FPN are scaled to 1/2, 1/4, 1/8, and so on in both the $x$ and $y$ directions. The scale gap between two adjacent layers of FPN is large and causes two objects with similar sizes to be predicted and categorized to different layers.  For example, two objects with the dimensions 32$\times$32 and 31$\times$31 are at different prediction maps in the FPN.  This scale truncation problem can be improved by adding a synthetic layer to this FPN.
\begin{figure}[ht]
\centering
\begin{tikzpicture}[scale=1.0]
\begin{axis}[
    y label style={at={(axis description cs:0.1,0.5)},anchor=south},
    xlabel=$Inference Time (ms)$,
    ylabel=$AP(\%)$,
    legend style={nodes={scale=0.5,transform shape}},
    legend pos=south east]

\addplot[smooth,mark=pentagon*,white!30!red] plot coordinates {
    (34.8189415,15.5)
    (35.99712023,16.4)
    (45.99816007,16.9)
};
\addlegendentry{MobV2-SFPN}

\addplot[smooth,mark=*,white!40!red] plot coordinates {
    (37.99392097,16.7)
    (52.00208008,17.6)
};
\addlegendentry{MobV2-SFPN-SOL}

\addplot[smooth,mark=pentagon*,white!30!orange] plot coordinates {
    (252.52525253,19.5)
    (255.7544757,19.7)
    (270.27027027,20.1)
};
\addlegendentry{VGG16-SFPN}

\addplot[smooth,mark=*,white!40!orange] plot coordinates {
    (261.78010471,20.2)
    (281.69014085,20.5)
};
\addlegendentry{VGG16-SFPN-SOL}

\addplot[smooth,mark=pentagon*,white!30!blue] plot coordinates {
    (125.94458438,20.4)
    (127.226463104,20.5)
    (136.054421769,20.8)
};
\addlegendentry{ResNet50-SFPN}

\addplot[smooth,mark=*,white!40!blue] plot coordinates {
    (131.92612137,20.9)
    (144.09221902,21.4)
};
\addlegendentry{ResNet50-SFPN-SOL}

\end{axis}
\end{tikzpicture}
\caption{SFPN with input size 224 on MS-COCO}
\end{figure}
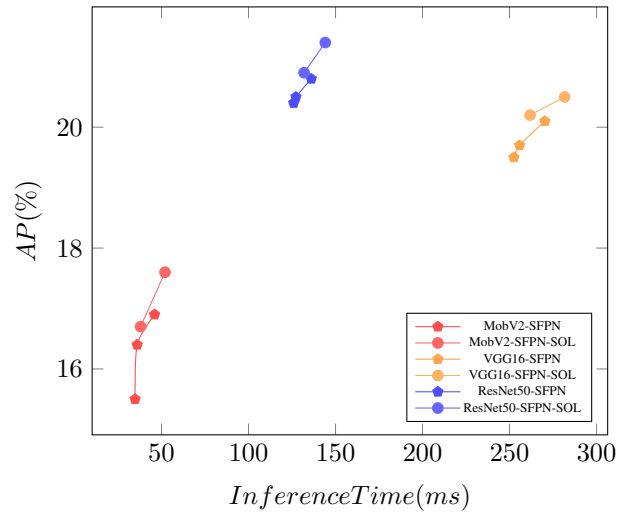

\begin{figure*}
\centering
\begin{tabular}{ccc}
\fbox{\includegraphics[width=5.cm]{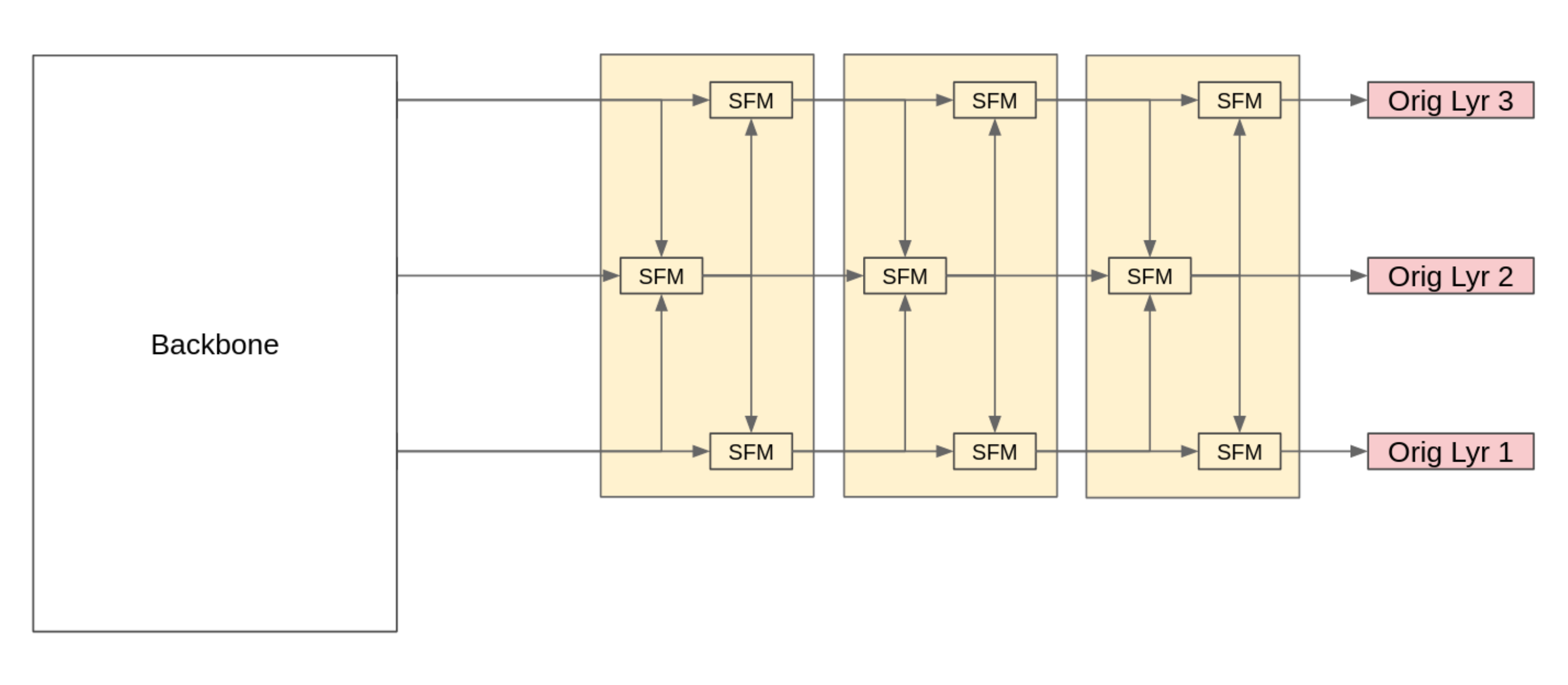}}&
\fbox{\includegraphics[width=5.3cm]{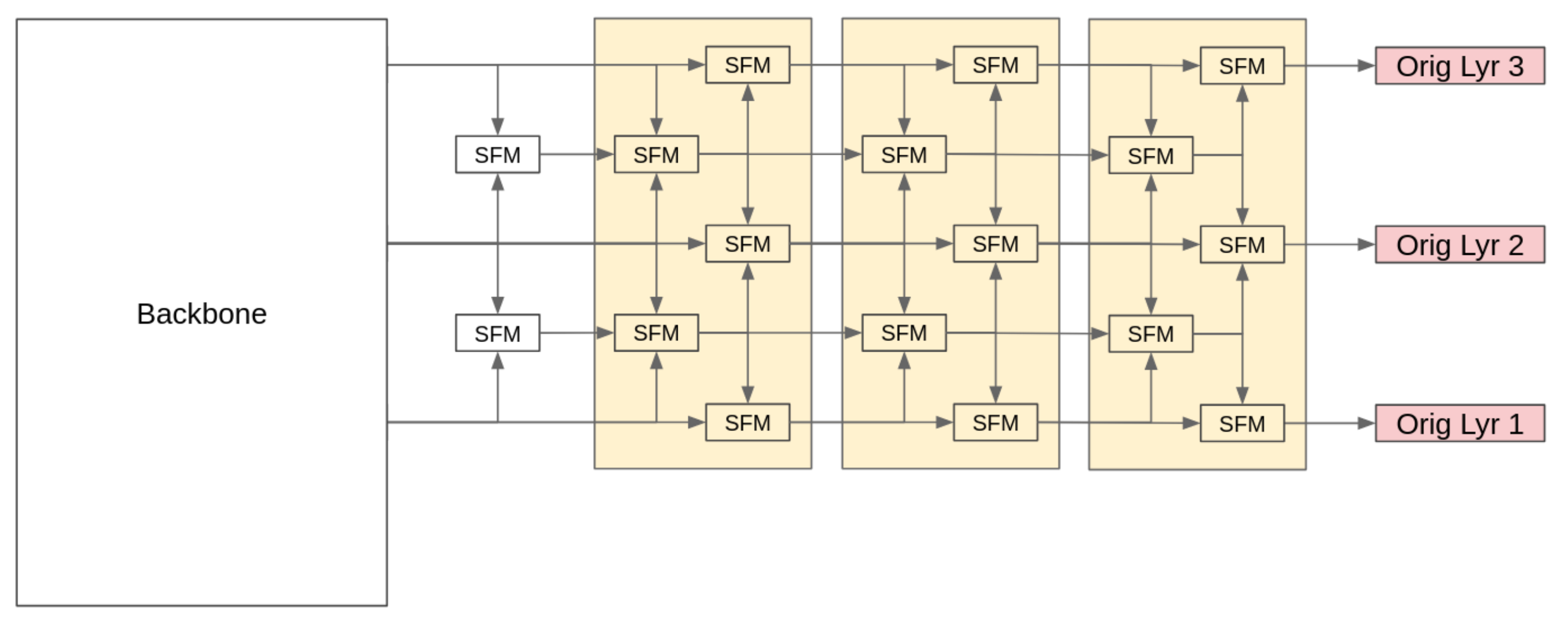}}&
\fbox{\includegraphics[width=5.45cm]{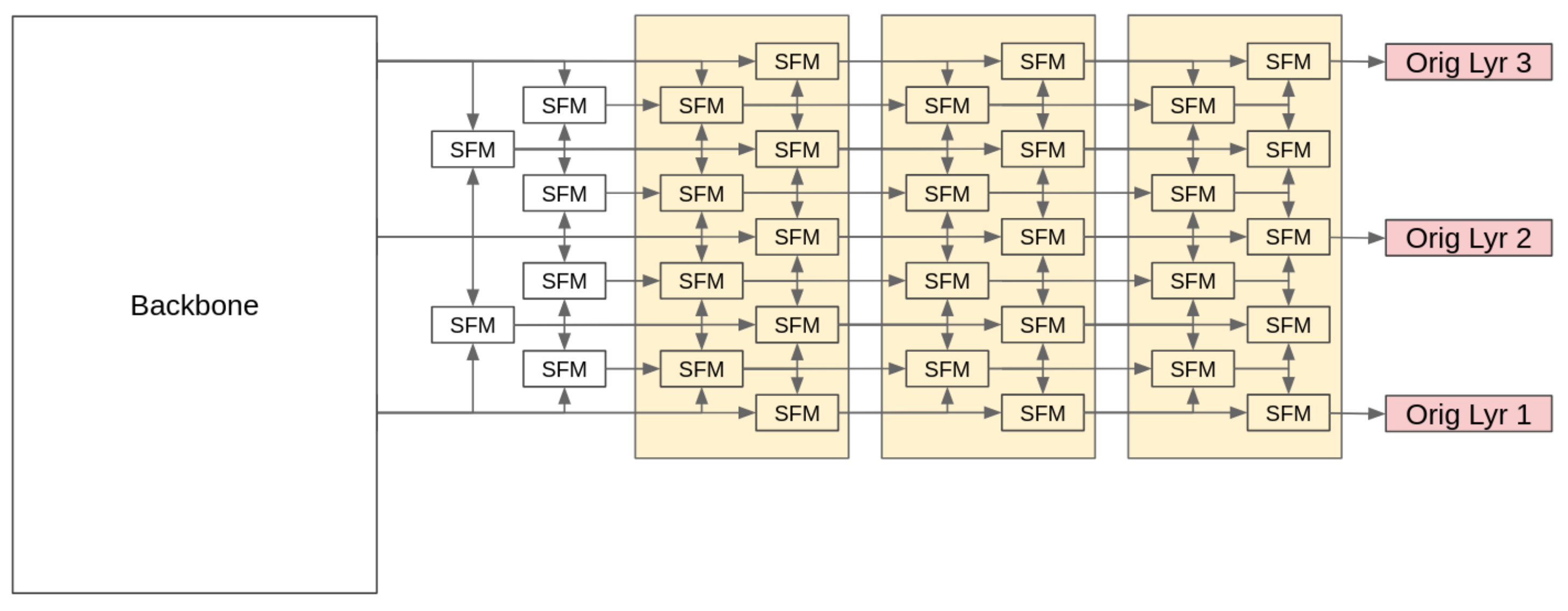}}\\
(a)&(b)&(c)
\end{tabular}
\caption{Synthetic Fusion Pyramid Network (SFPN): (a) SFPN-3, (b) SFPN-5,  and (c) SFPN-9. The yellow squares represent SFBs.}
\label{SFPN}
\end{figure*} 

This paper proposes a new SFPN (Syntethic  Fusion  Pyramid  Network) (see \myfigref{SFPN}) to make the density map scaled to 1/2, 1/3, 1/4, 1/6 and so on for reducing the effect of scaling truncation. We believe that the adding of middle-scale feature map can make the transition of different scales smoother for better object detection  especitally on a lightweight architecture. We build a baseline with only three outputs layers based on this idea, then gradually synthetically insert the middle-scale layers. The FPNs is called Syntethic Fusion Pyramid Network (SFPN). Then, the SFPN is integrated on VGG-16~\cite{vgg} and MobileNetV2~\cite{mobilenetv2}, respectively, and evaluated in the MS-COCO~\cite{mscoco} dataset. Unsurprisingly, we find that both large and lightweight backbones can benefit from the synthetic layers. The same scheme is applied again to separate the final output layers to more layers and surprisingly still outperforms the baseline.  Furthermore, we visualize the confidence maps of the SFPN to show how the synthetic layers can help the model fit the objects better. The synthetic layers can make the original layers retain more information for object representation. The findings confirm the performance gain of synthetic layers, and we hope that more variants of this promising method can be derived.

\section{RELATEDWORK}
\label{sec:format}
\subsection{Multi-Scale Prediction}
Object detection is a very active field in computer vision and can be organized into two categories based on their network architectures: {\em two-stage} proposal-driven and {\em one-stage} (single-shot) approaches. In general, two-stage methods such as Faster-RCNN~\cite{fasterrcnn} can achieve high detection accuracy but with longer computation time, while one-stage methods such as YOLO~\cite{yolov1,yolov2,yolov3,yolov4} run faster with inferior accuracy. We focus on the survey of one-stage object detectors. SSD~\cite{ssd} employed  in-network multiple feature maps for detecting objects with varyingshapes and  sizes. The multi-map design enabled SSD with better robustness over YOLOv1~\cite{yolov1}.
For  better  detection of  small  objects,  the  Feature  Pyramid  Network  (FPN)  \cite{fpn} based on FP can achieve higher detection accuracy for small objects. Now, FPN is widely used in SoTA detectors for detecting objects at different scales, where spatial and contextual features are extracted from the last layer of the top-down path for accurate object detection. This top-down aggregation is now a common practice for improving scale invariance in both two-stage and one-stage detectors.

\subsection{ Bi-directional FPN}

It is also well-known that the  top-down  pathway  in  FPN  cannot  preserve  accurate  object localization due to the shift-effect of pooling.  Bi-directional  FPN can recover lost information from shallow layers to improve {\bf small object detection} in several works \cite{LRF_Wang_2019_ICCV,Woo2019GatedBF,Wu2018SingleShotBP}.  For example,  PANet~\cite{panet} added the top-bottom direction after the bottom-up direction for significantly improving the expressive ability of FPN.
 A light-weight scratch network and a bi-directional network were constructed in \cite{LRF_Wang_2019_ICCV} to efficiently circulate both low- and high-level semantic information.  Inspirited by NAS-FPN \cite{nasfpn}, a BiFPN was proposed in \cite{efficientdet} to better detect small objects with higher efficiency. The recent YOLOv4 \cite{yolov4} modified the path aggregation method \cite{panet} by replacing the addition with concatenation to better detect small objects.  All the above methods prove this FPN with two directions outperforms the orginal FPN~\cite{fpn}. 

\begin{figure}
\centering
\fbox{\includegraphics[width=8.cm]{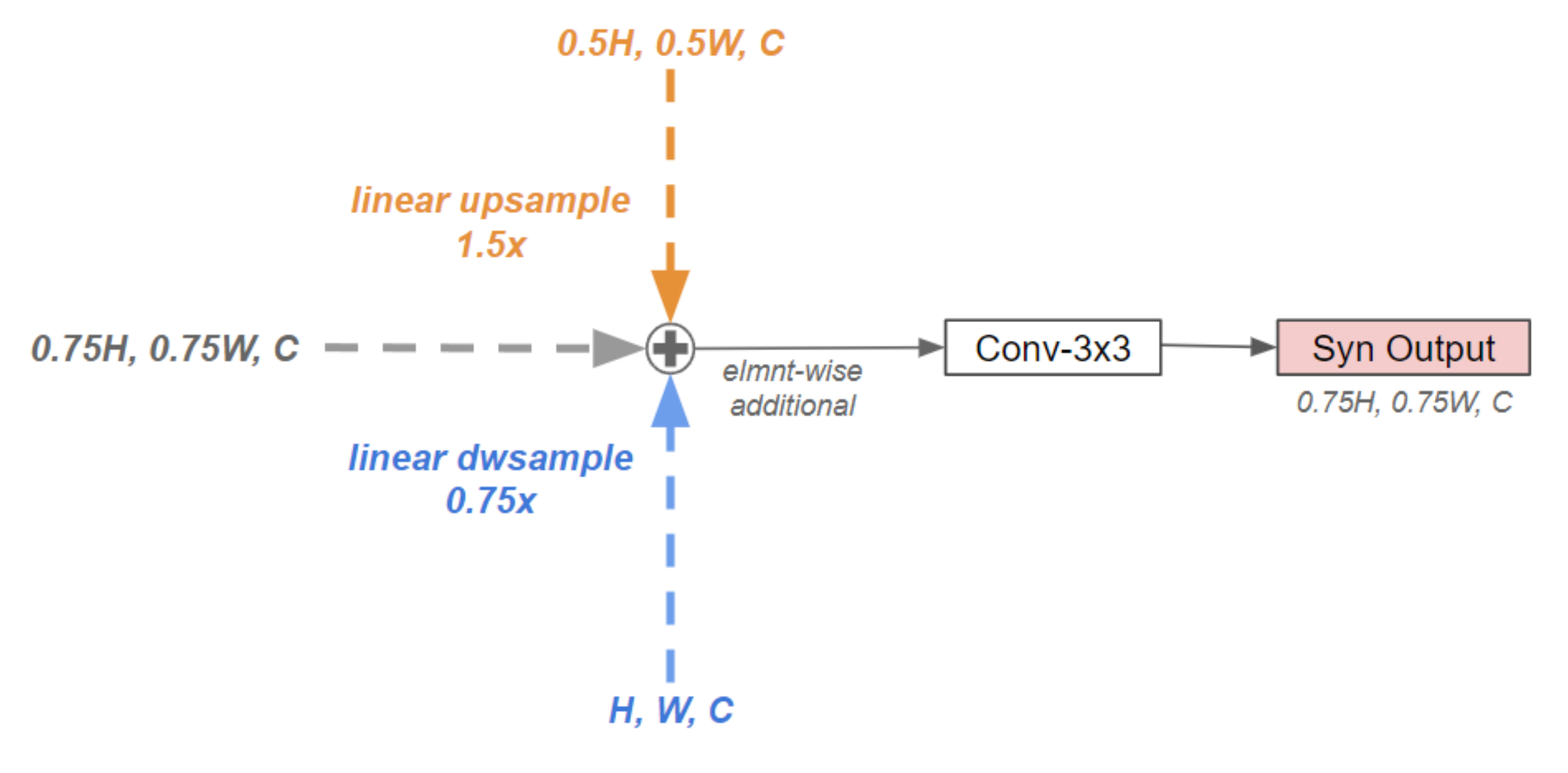}}\\
\caption{Architecture of Synthetic Fusion Module (SFM). The blue, orange and gray arrows represent the optional inputs.}
\label{sfm}
\end{figure}


\section{APPROACH}
\label{sec:pagestyle}
The pooling  operation (or convolution with stride 2)  used  in  CNN  backbone usually  down-samples the image dimension to half, and makes the densitymap scaled to 1/2, 1/4, 1/8, and so on in both the $x$ and $y$ directions. We believe that the scale gap is too big and causes the features fusion of layers not smooth. As shown in \myfigref{SFPN}, various synthetic layers between the original layers are created to make the prediction maps scaled to 1/2, 1/3, 1/4, 1/6 and so on, and thus a smoother scale space is provided for fitting the ground truth whose scale changes continuously. This section will describe how to generate these synthetic layers. After that, we visualize the SFPN to show how the synthetic layerscan help the model fit and detect objects better.

\subsection{Synthetic Fusion Module (SFM)}
FPN fuses the features of different layers along the top-down direction, and PANet finds that the bottom-up direction can also improve the performance. The later backbones such as NAS-FPN \cite{nasfpn} adopt the similar bifusion structure  to achieve better performance. This paper proposes a SFM (Synthetic Fusion Module) to generate various synthetic layers between the original layers so that the prediction maps are scaled to 1/2, 1/3, 1/4, 1/6 and so on. It contains three optional inputs, first linearly scaling inputs, then adding them pixel by pixel, and then fusion with a conv-3$\times$3. This module can synthesize the synthetic layers from the original layers or simply be used to fuse features. The architecture is shown in the \myfigref{sfm}.

\subsection{Synthetic Fusion Block (SFB)}
SFB is built from multiple SFMs. It divides the layers into two batches. First, the features are passed from the first batch of layers to the second batch of layers, then from the second batch of layers to the first batch of layers. The architecture is shown in the yellow block in \myfigref{SFPN}. In short, this idea can be treated as merging the features centrally and then radiating the features outward. SFB integrates the top-down and down-top directions in the same block. The stacking of multiple SFBs can perform multiple feature fusions for better performance. All models in the experiments are stacked with three SFBs.

\subsection{Synthetic Fusion Pyramid Network (SFPN)}
We call the FPN with SFB stacked three times as Synthetic Fusion Pyramid Network (SFPN), and we call the SFPN containing X output layers as SFPN-X.
\subsubsection{Build Baseline (SFPN-3)}
In order to verify that synthetic layers are an effective strategy, we use SFBs to construct an SFPN with only original layers, called SFPN-3. SFPN-3 maintains the original size as same as FPN. It adopts two-direction features fusion, which is precisely the same as other SFPNs that contain synthetic layers. In experiments, we use SFPN-3 as the baseline.
\subsubsection{SFPN-5 and SFPN-9}
SFM can generate synthetic layers, so adding several SFMs at the front of SFPN can generate several synthetic layers and then input the following three SFBs. The extension to five layers is called SFPN-5, and the extension to nine layers is called SFPN-9. These two networks have added three and six synthetic layers, respectively, which are the main models to verify the effectiveness of synthetic layers.
\subsubsection{SFPN with Synthetic Output Layers}
\begin{figure*}
\centering
\begin{tabular}{cc}
\fbox{\includegraphics[width=7cm]{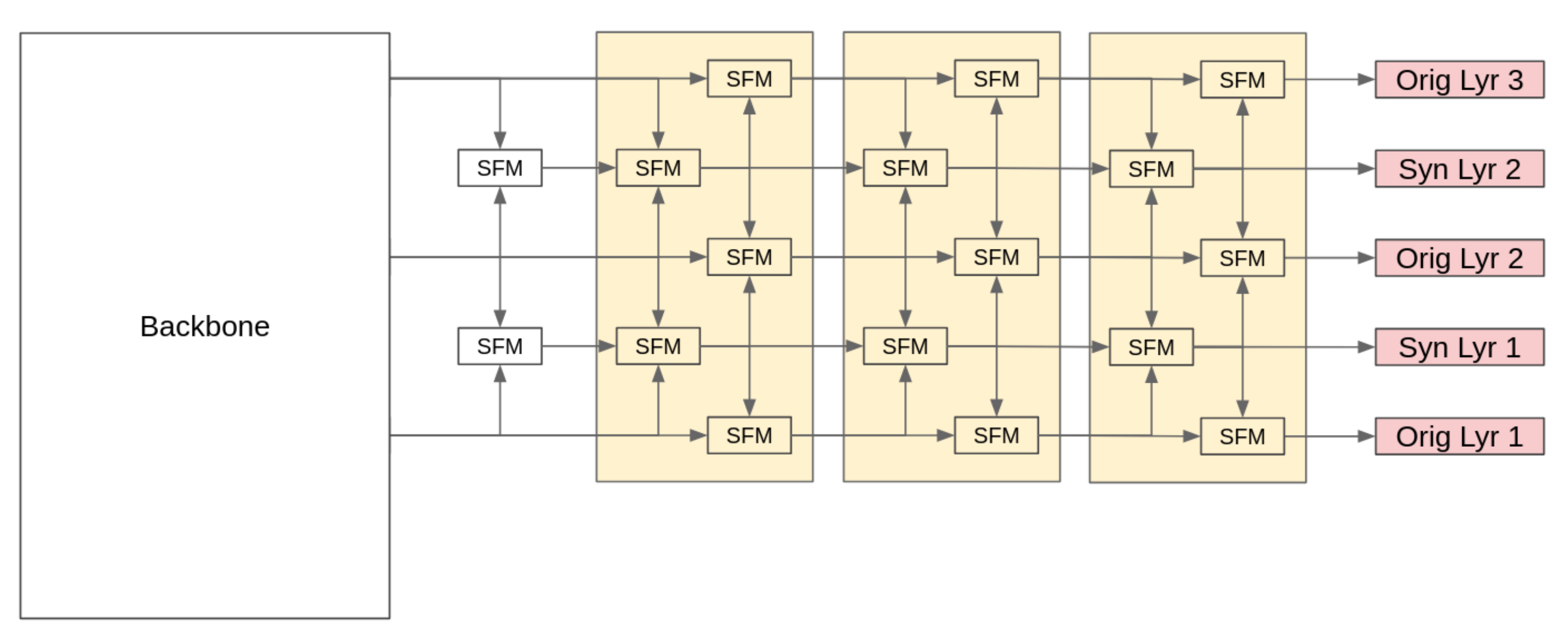}}&
\fbox{\includegraphics[width=7.2cm]{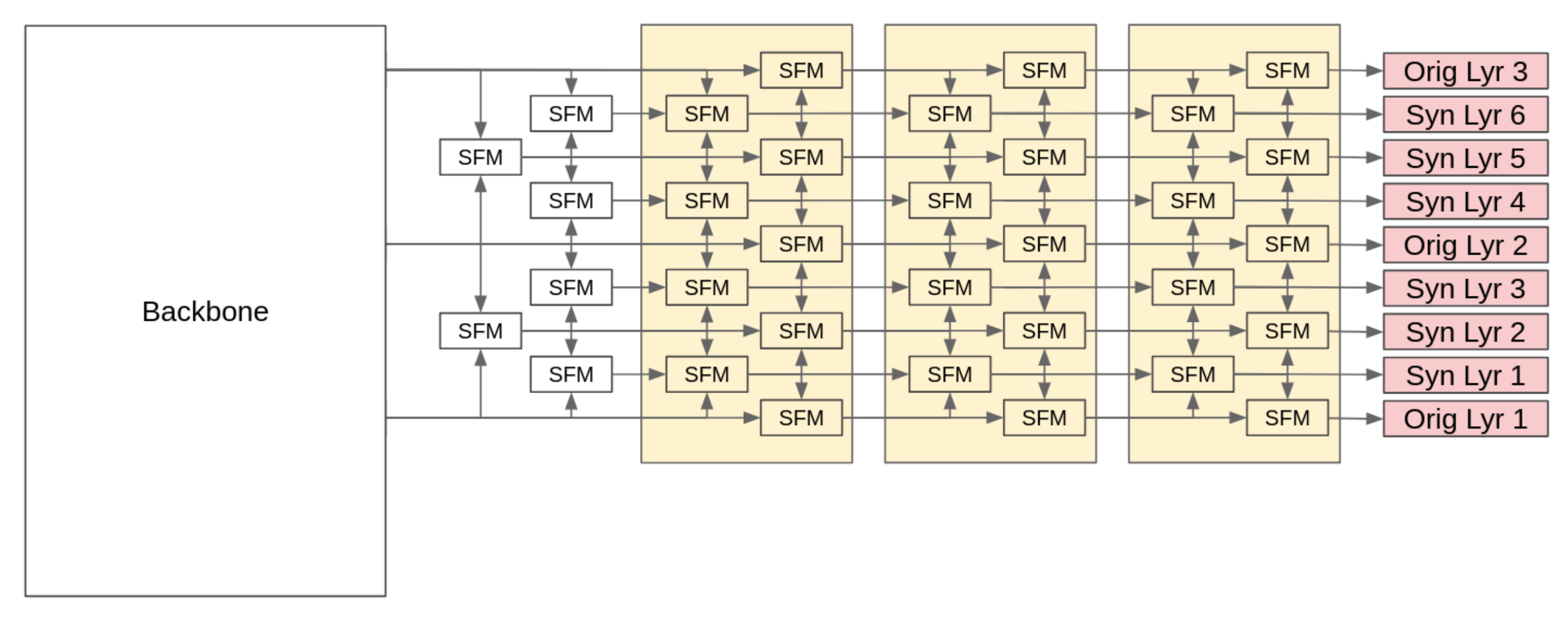}}\\
(a)&(b)
\end{tabular}
\caption{Synthetic Fusion Pyramid Network with synthetic outputs layers: (a) SFPN-5-SOL, (b) SFPN-9-SOL. The yellow squares represent SFBs.}
\label{SFPNwsol}
\end{figure*}

The proposed synthetic layers make the scales of fetaures more continuous and make them transfer smoother during the features fusion stage. In order to further explore the ability of this component, we add the synthetic output layers of the SFPN-5 and SFPN-9. These networks are recorded as SFPN-5-SOL and SFPN-9-SOL. Please refer to \myfigref{SFPNwsol} for details.
\begin{figure}
\centering
\fbox{\includegraphics[width=7.cm]{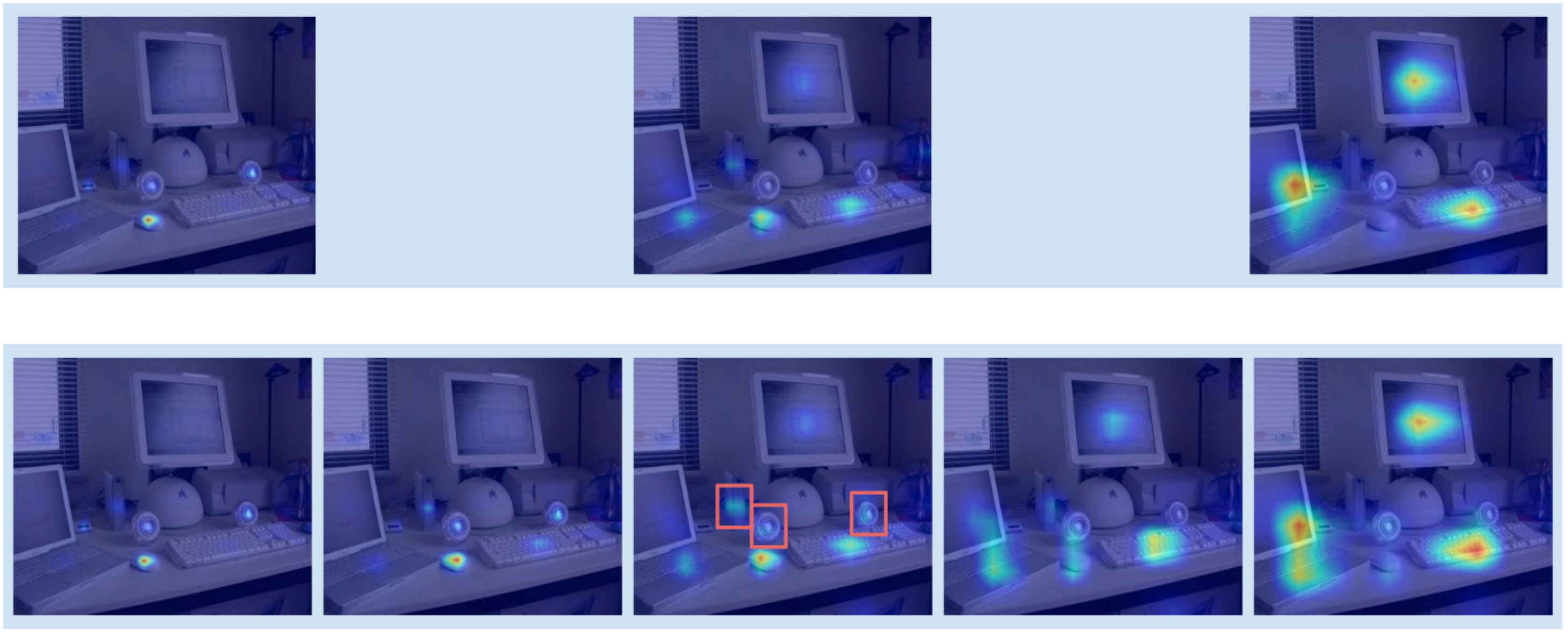}}\\
\caption{The synthetic layers make the original layers reduce the feature loss. The top three frames are the confidence maps output by SFPN-3, and the bottom three frames are the confidence maps output by SFPN-5-SOL.}
\label{moresmoother}
\end{figure}
\subsection{Naive Anchors for SOL}
We connect SFPN to the YOLO head. In the YOLO architecture, the generating method of anchors and the allocation strategy affect the object detector's performance significantly. YOLO uses k-means to find k prior boxes in the training set as anchors. CSL-YOLO found that when the number of output layers increases, k-means will generate many anchors that do not fit the scale of the output layers. We adopted a straightforward generating method of anchors to remove this significant interference factor. We use priority boxes with a ratio of 1x, 2x, and 4x as anchors on every pixel of the output feature maps. This method enables the output of three layers, five layers, and nine layers to obtain scale-fitting and consistent anchors, proving that the performance gains obtained are derived from synthetic output layers.
\begin{figure}
\centering
\fbox{\includegraphics[width=7cm]{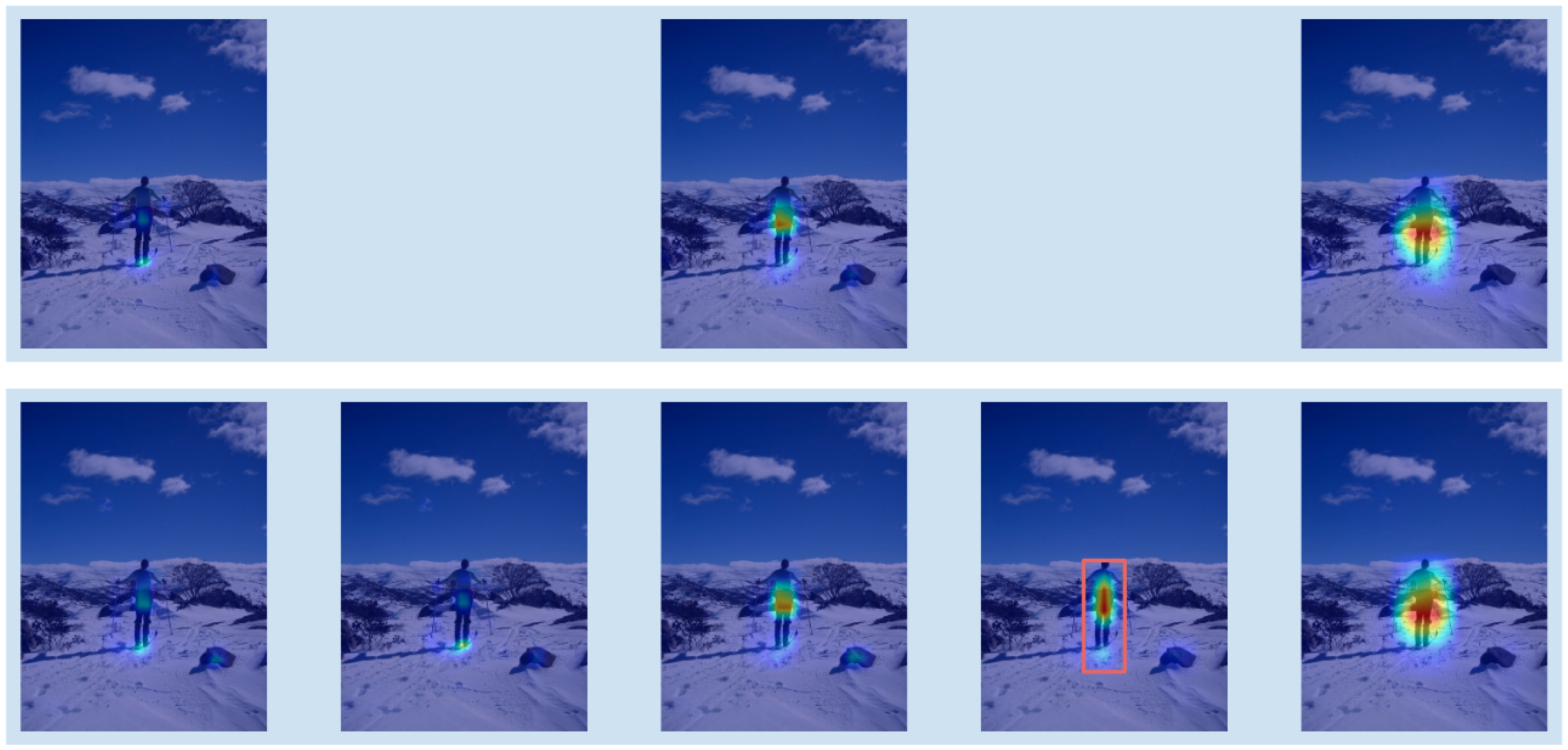}}\\
\caption{Synthetic layers fit some objects better than original layers. The top three frames are the confidence maps output by SFPN-3, and the bottom three frames are the confidence maps output by SFPN-5-SOL.}
\label{morefitting}
\end{figure}
\section{EXPERIMENTS}
\label{sec:typestyle}
We evaluate SFPN using MS-COCO, a public and representative object detection dataset. In order to ensure the consistency of the experiments, we adopt fixed hyperparameter settings, visualize the confidence map to analyze the effect of synthetic layers, and finally provide detailed experimental results to prove that SFPN is an efficacious method.

\subsection{Model Setting}
We use VGG-16 and MobileNetV2 as backbones with limited computing resources, both of which are pre-trained on ImageNet\cite{imagenet}. These two models represent classic considerable models and miniature models. We use them to evaluate the performance of SFPN on MS-COCO. Since the sampled channel of backbone is not the same, we fixed the channel at 112 with conv-3x3 before the neck (SFPN) to facilitate the element-wise additional feature fusion adopted by SFPN. We fixedly stacked three SFBs in SFPN and connected the YOLO head to build an object detection model. The same augmentation strategy, learning schedule, etc......, are used during training.
\begin{table}[h]
\caption{Comparison of SFPN-5 and SFPN-9 with the baseline (SFPN-3) on MS-COCO. Note that the FPS is derived from the I7-6600U.}
\begin{center}
\setlength{\tabcolsep}{0.9mm}{\small
\begin{tabular}{|c|c|c|c|c|c|c|}
\hline
Model & Size & Params & FPS & AP & AP50 & AP75\\
\hline
MobV2-3 & 224 & 2.7M & 28.72 & 15.5 & 30.2 & 14.7\\
MobV2-5 & 224 & {3.6M} & 27.78 & {16.4} & {31.4} & {15.5}\\
MobV2-9 & 224 & {5.4M} & 21.74 & {16.9} & {32.0} & {16.3}\\
\textbf{MobV2-5-SOL} & \textbf{224} & \textbf{3.6M} & 26.32 & \textbf{16.7} & \textbf{31.9} & \textbf{15.9}\\
\textbf{MobV2-9-SOL} & \textbf{224} & \textbf{5.4M} & 19.23 & \textbf{17.6} & \textbf{33.2} & \textbf{17.3}\\

MobV2-3 & 320 & 2.7M & 17.86 & 19.9 & 37.8 & 19.1\\
{MobV2-5} & {320} & {3.6M} & 14.71 & {20.4} & {38.2} & {19.9}\\
{MobV2-9} & {320} & {5.4M} & 11.11 & {20.5} & {38.4} & {19.8}\\
\textbf{MobV2-5-SOL} & \textbf{320} & \textbf{3.6M} & 13.51 & \textbf{20.6} & \textbf{38.4} & \textbf{20.2}\\
\textbf{MobV2-9-SOL} & \textbf{320} & \textbf{5.4M} & 9.62 & \textbf{21.1} & \textbf{38.8} & \textbf{20.7}\\
\hline
VGG16-3 & 224 & 17.1M & 3.96 & 19.5 & 36.5 & 18.8\\
{VGG16-5} & {224} & {18.0M} & 3.91 & {19.7} & {36.6} & {19.5}\\
{VGG16-9} & {224} & {19.8M} & 3.70 & {20.1} & {36.9} & {20.0}\\
\textbf{VGG16-5-SOL} & \textbf{224} & \textbf{18.0M} & 3.82 & \textbf{20.2} & \textbf{37.2} & \textbf{20.1}\\
\textbf{VGG16-9-SOL} & \textbf{224} & \textbf{19.8M} & 3.55 & \textbf{20.5} & \textbf{37.3} & \textbf{20.5}\\

VGG16-3 & 320 & 17.1M & 1.95 & 20.5 & 38.7 & 19.7\\
{VGG16-5} & {320} & {18.0M} & 1.89 & {20.7} & {38.7} & {20.4}\\
{VGG16-9} & {320} & {19.8M} & 1.82 & {22.6} & {41.0} & {22.5}\\
\textbf{VGG16-5-SOL} & \textbf{320} & \textbf{18.0M} & 1.89 & \textbf{21.3} & \textbf{39.4} & \textbf{20.9}\\
\textbf{VGG16-9-SOL} & \textbf{320} & \textbf{19.8M} & 1.77 & \textbf{23.1} & \textbf{41.7} & \textbf{23.2}\\

ResNet50-3 & 224 & 28.3M & 7.94 & 20.4 & 36.2 & 20.7\\
{ResNet50-5} & {224} & {29.2M} & 7.86 & {20.4} & {36.3} & {20.8}\\
{ResNet50-9} & {224} & {31.0M} & 7.35 & {20.8} & {36.8} & {21.2}\\
\textbf{ResNet50-5-SOL} & \textbf{224} & \textbf{29.2M} & 7.58 & \textbf{20.9} & \textbf{36.8} & \textbf{21.4}\\
\textbf{ResNet50-9-SOL} & \textbf{224} & \textbf{31.0M} & 6.94 & \textbf{21.4} & \textbf{37.5} & \textbf{22.1}\\

\hline
\end{tabular}}
\end{center}
\label{mainresult}
\end{table}
\subsection{More Suitable Feature}
Adding more synthetic layers increases the number of output layers of SFPN from three to five and finally to nine. In order to further explore the cascading benefits brought by these synthetic layers, we visualize the confidence map output by SFPN-5 and draw it on the original image above, as shown in \myfigref{moresmoother}. The scale of the synthetic layers is between the upper and lower original layers, which allows the original layers to transfer features more smoothly, thereby reducing the loss of features. On the other hand, the size of some objects is more in line with the feature description of synthetic layers than original layers, which enables the model to obtain more suitable object representation capabilities, as shown in \myfigref{morefitting}. In general, this method enables original layers to reduce feature loss and consider more objects of different sizes, and the new synthetic layers can also be more adaptable to different object sizes to predict more appropriate bounding boxes.

\subsection{Results}
The final experiment is shown in \mytabref{mainresult}. If the visual confidence map only provides intuitive evidence, then the performance on MS-COCO is the direct evidence. When the backbone is VGG-16, more synthetic layers can get a higher AP, whether the input image size is 224 or 320. On the other hand, when the backbone is MobileNetV2, the AP increase is more pronounced. It can be said that the feature capture ability of SFPN is more prominent on weaker small models.

\subsection{More Output Layers}
We add the last synthetic layers as the output layer on the converged SFPN-5 and SFPN-9 and predict the test set with new output layers. We want to use this experiment to evaluate how synthetic output layers with continuous scales can significantly improve performance. \mytabref{mainresult} shows the results of this experiment. Although the modified SFPN-5-SOLand SFPN-9-SOL lost some FPS, they surpassed the baseline more, confirming that synthetic layers played an essential role not only in the feature fusion stage but also in the output stage.

\vfill\pagebreak
\bibliographystyle{IEEEbib}
\bibliography{refs}

\end{document}